\documentclass[letterpaper,10pt,conference]{ieeeconf}

\IEEEoverridecommandlockouts
\usepackage{amsmath,amssymb,bm}
\usepackage{booktabs}
\usepackage{array}
\usepackage{cite}
\usepackage{graphicx}
\usepackage{microtype}
\usepackage{xurl}
\usepackage{booktabs,tabularx,amsmath}
\usepackage[T1]{fontenc}

\usepackage{tikz}
\usetikzlibrary{arrows.meta,positioning}

\usepackage[hidelinks]{hyperref}

\newcolumntype{L}[1]{>{\raggedright\arraybackslash}p{#1}}

\title{\LARGE \bf
IL-ACT: Imitation Learning with Adaptive Cartesian Tracking Control for a 30-ton Excavator}

\author{{Mehdi Heydari Shahna, Seihun Kim, Soyi Jung, Soohyun Park, Jouni Mattila, Joongheon Kim}
\thanks{This work was supported by the Korean Ministry of Trade, Industry and Energy (MOTIE) under Grant RS-2024-00442168, through the Mechanical Equipment Industry Technology Development Program administered by KEIT.
M. H. Shahna and J. Mattila are with the Faculty of Engineering and Natural Sciences, Tampere University, Tampere, Finland; S. Kim and J. Kim are with the Department of Defense Convergence Technology and the Department of Electrical and Computer Engineering, respectively, at Korea University, Seoul, Republic of Korea; S. Jung is with the Department of Electrical and Computer Engineering, Ajou University, Suwon, Republic of Korea; and S. Park is with the Division of Computer Science, Sookmyung Women’s University, Seoul, Republic of Korea. (Corresponding author e-mail: mehdi.heydarishahna@tuni.fi)}
}

\begin{document}
\maketitle
\thispagestyle{empty}
\pagestyle{empty}

\begin{abstract}
Autonomous excavator control is challenged by coupled kinematics,
actuation lag, and uncertainty. We propose imitation learning and adaptive Cartesian tracking (IL-ACT), a novel motion control framework for a
30-ton-class excavator. An anchored, 14-input imitation policy
pretrained on operator demonstrations generates nominal joint
rates; adaptive Cartesian feedback and gated gain/bias estimation
correct these commands before a stopping-distance governor
constrains joint-reference generation.
Simscape evaluation covers 100 sequential goals and spiral,
figure-eight, and rounded-raster tracking, including 88 additional
runs across three training seeds, two initializations, and speeds,
under hydraulic response and sensing conditions. Compared with
Teacher+ACT, IL-ACT completes all goals with shorter duration
and lower terminal errors under both response conditions.
Telemetry-initialized IL-ACT lowers RMSE in all 24 figure-eight
and rounded-raster seed comparisons and lowers additional-load
spiral mean RMSE by approximately $29\%$.
Original spiral RMSE also improves over IL-only and PID.
Under a shared sensor-noise realization, telemetry-initialized
IL-ACT achieves $27.67\%$ lower mean RMSE than Teacher+ACT;
enabling estimation reduces mean RMSE by $22.44\%$ relative
to the frozen estimator. Pretrained-weight effects
remain mixed, and the original teacher comparison exhibits a
spiral RMSE--maximum-error tradeoff. Analysis establishes bounded
adaptive states and Cartesian feedback, with reference admissibility
conditional on governor feasibility.
\end{abstract}

\section{Introduction}

\begin{figure}[h!]
\hspace*{-0.0cm} 
\centering
\scalebox{1}{\includegraphics[trim={0cm 0.0cm 0.0cm 0cm},clip,width=\columnwidth]{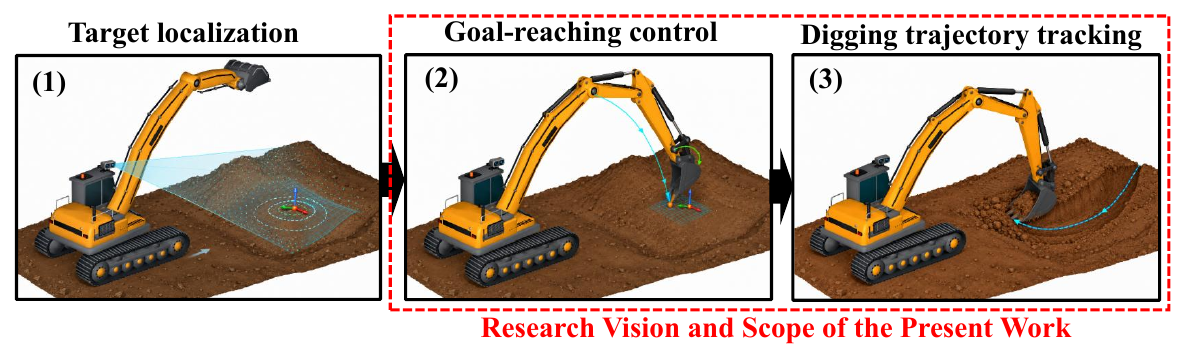}}
\caption{Stages of autonomous excavation and scope of the present work.}
\label{fff}
\end{figure}

Autonomous excavation requires target localization and machine positioning,
goal-reaching control, and trajectory tracking (Fig.~\ref{fff}). This work designs a new motion control framework, addressing the latter two stages for a 30-ton-class excavator with coupled
kinematics and hydraulic actuation. Previous systems have demonstrated
trajectory optimization, constrained planning, learned inverse control, and
integrated excavation~\cite{Yang2021ExcavationTrajectory,
Lee2021ExcavatorMPC,Lee2022DataDrivenExcavator,Jang2025IntegratedExcavation}.

Imitation learning (IL) exploits expert demonstrations and has been combined
with model-based trajectory generation and learned closed-loop excavator
dynamics~\cite{Guo2022ImitationExcavator,Zou2025FeedbackACT,Zou2025EfficientTrack,heo2026lightweight}.
Offline robot-learning studies emphasize data quality, temporal context,
distribution shift, and reliable evaluation~\cite{Mandlekar2022Offline,
Zhou2025MTIL,Mehta2025StableBC,Vincent2024Generalizable}. Causal confusion
further motivates observation design consistent with deployment-time
measurements~\cite{DeHaan2019Causal}.

Recent robotics research has combined adaptive control with neural networks to improve tracking accuracy and robustness under uncertain dynamics and changing operating conditions. Kong et al.~\cite{kong2024adaptive} developed an adaptive neural controller that compensates for unknown robot dynamics and actuator saturation, establishing fixed-time convergence of tracking errors to a neighborhood of zero and validating the approach through simulations and physical experiments. For magnetic micromanipulation, Jia et al.~\cite{jia2024efficient} integrated offline neural model learning with online weight adaptation and constrained optimal control, enabling experimentally demonstrated non-contact trajectory tracking and navigation in cluttered environments. Extending adaptation to changing payloads, Li et al.~\cite{li2025physics} combined online payload identification with physics-informed neural networks and predictive control. Their work reported approximately 35\% higher tracking precision than their previous adaptive controller in quadruped experiments with payloads ranging from 25 to 100~kg. More recently, Han et al.~\cite{han2026tracking} combined adaptive radial basis function neural networks with local model approximation and particle swarm optimization, reporting reduced tracking errors in simulations of an ABB IRB1600 industrial manipulator. Collectively, these studies suggest that neural approximation and online adaptation provide complementary mechanisms for improving robotic control performance, although the differing platforms, baselines, and validation methods limit direct comparisons between their reported gains. Adaptive control and other control-theoretic methods can strengthen neural-network-based reinforcement learning (RL) and imitation learning by introducing explicit mechanisms for uncertainty compensation and error convergence \cite{shahna2025anti, shahna2024exponential}. Sung et al.~\cite{sung2024robust} augmented model-based RL with $\mathcal{L}_1$ adaptive control, using online uncertainty estimation and compensation to improve robustness and sample efficiency in simulated control tasks. Zhao et al.~\cite{zhao2023stable} incorporated control Lyapunov and barrier functions into actor-critic learning, expressing stability and safety requirements as optimization constraints. For imitation learning, Abyaneh et al.~\cite{abyaneh2025contractive} designed neural dynamical policies with contraction guarantees, ensuring convergence between generated trajectories and improving recovery from unfamiliar states. Such mathematical structure makes key behavioral properties analyzable even when the neural network itself remains difficult to interpret \cite{shahna2025anti}.

These architectures
motivate integrating a learned excavator policy with feedback that
compensates for uncertain joint response. Hence, for a 30-ton excavator,
we design IL with an adaptive Cartesian tracking (ACT) framework, denoted
IL-ACT. It coordinates an anchored joint-rate policy with nominal
response prediction, excitation- and intervention-gated gain/bias
estimation, and a shared stopping-distance governor.
The main contributions are:

1) \textit{Anchored imitation-policy design}. Kinematic teacher
supervision and dataset aggregation refine a coordinated four-joint
policy with causal observations and an exact zero-action anchor.
Telemetry-initialized and random-weight variants are evaluated.

2) \textit{Adaptive tracking with constrained reference generation}.
Cartesian feedback and gated projected gain/bias estimation share
a stopping-distance governor. The analysis establishes boundedness
and reference admissibility conditional on governor feasibility.

3) \textit{Comparative simulation evidence across tasks and trajectories}.
Goal regulation and spiral tracking include IL-only and tuned PID
baselines. An additional 88-run study compares Teacher+ACT and
IL-ACT across figure-eight, rounded-raster, and perturbed spiral
tracking, with three training seeds, two initializations, and
estimator ablations.

The comprehensive companion technical report documents execution,
training, response modeling, and baseline tuning (S1--S4), together
with ablations, detailed results, and validation checks (S5--S8).
Its proposed physical-machine protocol (S9) covers calibration,
actuator integration, point reaching and digging, safety supervision,
staged commissioning, independent measurement, and comparative
evaluation. The report and task video are available at
\nolinkurl{https://anonymous.4open.science/r/icra2027-supplement-6263}, on Anonymous GitHub.

\begin{figure*}[!t]
\centering
\includegraphics[
width=0.75\textwidth,
height=0.350\textheight,
keepaspectratio
]{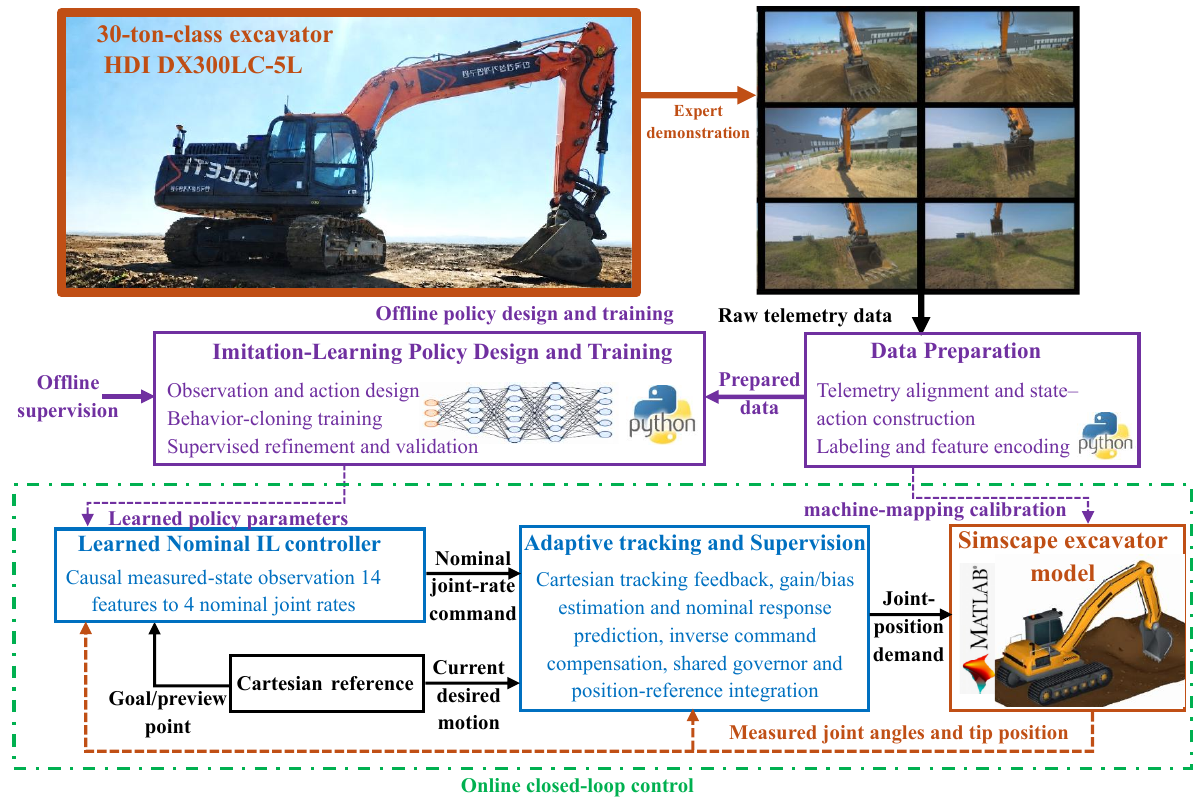}
\caption{IL-ACT architecture: offline policy training and validation, followed by online
Cartesian feedback, gain/bias adaptation, and command supervision.}
\label{marg}
\end{figure*}

\section{Preliminaries and Problem Formulation}
\label{sec:problem}

The machine is a 30-ton-class hydraulic excavator with four controlled
joints: swing, boom, arm, and bucket. Let
$\mathbf q=[q_1,q_2,q_3,q_4]^{\mathsf T}$ denote the model-aligned joint
angles in degrees, and let $\boldsymbol\theta=\kappa\mathbf q$, where
$\kappa=\pi/180$, denote the corresponding angles in radians.
Swing is an actively controlled joint in both goal regulation and timed
trajectory tracking. The controlled task is the three-dimensional
position of the center-reference bucket tip; bucket orientation is not
an independently tracked task variable.
The model-base origin lies on the swing axis at boom-pivot elevation,
with its vertical axis upward. Boom angle is measured from the swung
horizontal axis; arm and bucket angles are relative to their preceding
links. The radial swing-to-boom offset is $a_1=0.120$ m; the boom
and arm pivot-to-pivot lengths are $a_2=6.245$ and $a_3=3.113$ m,
and the bucket pivot-to-tip length is $a_4=1.910$ m.
All Cartesian references use this frame.
The conventional DH parameters and execution conventions are given in
Sec. S1.

For zero lateral bucket offset, define
\begin{equation}
\begin{aligned}
r={}&a_1+a_2\cos\theta_2+a_3\cos(\theta_2+\theta_3)\\
&+a_4\cos(\theta_2+\theta_3+\theta_4).
\end{aligned}
\end{equation}
The tip coordinates in metres are
\begin{equation}
\begin{aligned}
x&=r\cos\theta_1,\qquad y=r\sin\theta_1,\\
z&=a_2\sin\theta_2+a_3\sin(\theta_2+\theta_3)\\
&\quad+a_4\sin(\theta_2+\theta_3+\theta_4).
\end{aligned}
\label{eq:bucket_tip_fk}
\end{equation}
To match the controller's numerical units, define
\begin{equation}
\mathbf p=1000[x,y,z]^{\mathsf T},\qquad
\mathbf J(\mathbf q)=\frac{\partial\mathbf p}{\partial\mathbf q}
\in\mathbb R^{3\times4}.
\label{eq:cartesian_task}
\end{equation}
Thus, Cartesian positions and velocities use millimetres and
$\mathrm{mm\,s^{-1}}$, whereas joint rates use
$\mathrm{deg\,s^{-1}}$. The Jacobian includes the degree-to-radian
factor $\kappa$.

At period $T_c=0.1$ s, the controller receives measured joint angles
$\mathbf q_k$, measured tip position $\mathbf p_k$, the current desired
position $\mathbf p_{d,k}$ and velocity $\mathbf v_{d,k}$, and a
conditioning point $\mathbf p^{\mathrm{pr}}_k$ for the learned policy.
For goal regulation, $\mathbf p^{\mathrm{pr}}_k=\mathbf p_{d,k}$ is the
current fixed goal and $\mathbf v_{d,k}=\mathbf0$.
For timed tracking, the learned policy receives a preview point, while
feedback uses the current reference position and velocity. The timed
reference advances independently of tracking error; it is not paused
at intermediate points.

The controller generates a four-joint rate command $\mathbf u_k$ and
integrates it into a position-demand register $\mathbf c_k$:
\begin{equation}
\mathbf c_{k+1}=\mathbf c_k+T_c\mathbf u_k.
\label{eq:reference_integrator}
\end{equation}
The current register value $\mathbf c_k$ is emitted to the plant during
the current interval. The register and measured joint positions are
distinct. Initialization uses
$\mathbf c_0=[0,30,-100,-20]^{\mathsf T}$ degrees and requires measured
joints to agree with this initial demand within $10^{-3}$ degrees.
The control objective is Cartesian tracking with admissible generated
joint references. The governor constrains the position-demand register,
and measured bucket-tip motion determines Cartesian tracking performance.

\section{IL Policy Design and Training}
\label{sec:il_provider}
The dataset counts and checkpoint statistics in this section
describe the original telemetry-initialized benchmark policy.
The additional seed and initialization study is reported in
Section~\ref{subsec:expanded_tracking}.
The source telemetry corpus contains operator demonstrations from
$15$ recording dates, partitioned by date into seven training,
two validation, three test, and three stress-test dates.
We adapt the behavior-cloning approach in~\cite{heo2026lightweight}
to goal-conditioned four-joint excavator control.
For this original policy, all four network layers inherit weights
and biases from a telemetry-trained policy refined using telemetry replay
and offline kinematic supervision. The inherited input means and
scales $(\boldsymbol\mu_x,\boldsymbol\sigma_x)$ and affine output
transformation $(\mathbf A_y,\mathbf b_y)$ remain fixed.
The present refinement optimizes all shared network weights and
biases through both evaluations of the anchored policy in
\eqref{eq:anchored_policy}.
The hydraulic-response parameters are prescribed in Sec. S3.
The initial supervised dataset contains $120000$ accepted examples.
A kinematic teacher reconstructs the current configuration and Cartesian
goal from the observation without receiving a goal-joint witness.
It selects a bounded elbow-down IK posture and blends a bounded posture
guide with a box-constrained least-squares Cartesian-velocity command
as the goal approaches. Labels are joint-rate commands before the
stateful governor; measured-rate features do not enter the teacher law.
Three dataset-aggregation rounds each collect $96$ learner-driven
analytical rollouts, labeling visited states including unsuccessful
prefixes. Dataset sizes are $120000$, $135990$, $150677$, and
$164585$, with $60$ epochs per stage. The teacher supplies online
nominal commands only in the matched Teacher+ACT comparison.
Its exact law, sampling distributions, and collection procedure are
specified in Sec. S2.
For minibatch $\mathcal B$, define
$\mathbf r_n=\pi_\Theta(\mathbf x_n)-\mathbf u_n^{\rm E}$,
where $\pi_\Theta$ is the anchored policy in
\eqref{eq:anchored_policy} and $\mathbf u_n^{\rm E}$ is the teacher
label. Here $\widetilde{\mathbf q}_n$ is the joint configuration reconstructed
from the pose features of $\mathbf x_n$.
With $\mathbf J_n=\mathbf J(\widetilde{\mathbf q}_n)$ in
$\mathrm{mm\,deg^{-1}}$, the dimensionless training loss is
\begin{equation}
\mathcal L_{\mathcal B}(\Theta)=
\frac{1}{|\mathcal B|}\sum_{n\in\mathcal B}w_n
\left[
\frac{\|\mathbf D_v^{-1}\mathbf r_n\|_2^2}{4}
+\frac{0.2}{3}\left\|\frac{\mathbf J_n\mathbf r_n}{60}\right\|_2^2
\right],
\label{eq:complete_training_loss}
\end{equation}
where $\mathbf D_v=\operatorname{diag}(0.6,0.4,0.6,0.8)$ uses
$\mathrm{deg\,s^{-1}}$, the Cartesian normalization is
$60\,\mathrm{mm\,s^{-1}}$, and
$w_n=1+3\exp[-\max(x_{n,13},0)/(100\,\mathrm{mm})]$.
The loss averages over examples, not the sum of weights, and
compares predicted and teacher Cartesian velocities.
Its definition is fixed while dataset aggregation changes the
empirical training distribution.
Each stage uses AdamW. Checkpoint selection maximizes success on
the same 32 validation rollouts and then minimizes mean terminal error.
The selected fourth-stage checkpoint is evaluated on 64 separate test
cases. Optimizer settings, seeds, and complete training details are
provided in Sec. S2.

\section{Control Design}
\label{sec:control}

For each IL-only versus IL-ACT comparison, both modes use the
same learned policy, observations, tasks, and command governor.
IL-only disables Cartesian correction and gain/bias adaptation;
IL-ACT enables both upstream of the governor. 

The nominal plant response uses two cascaded $0.1$ s position-input
filters. The disturbed response adds valve, friction,
and supply/load effects
\cite{RudermanHydraulic2017,shahna2025anti,sadsadshahna2025robust},
described in Section~\ref{sec:evaluation} and Sec. S3.

\subsection{Design of the Goal-Conditioned Imitation Policy}
\label{subsec:nominal_il}

The observer unwraps swing using shortest angular increments and applies
causal exponential filtering to the measured joints and tip position.
With $\alpha_o=1-\exp(-T_c/0.15)$,
\begin{equation}
\begin{aligned}
\widehat{\mathbf q}_k&=(1-\alpha_o)\widehat{\mathbf q}_{k-1}
 +\alpha_o\mathbf q^{\mathrm{uw}}_k,\\
\widehat{\mathbf p}_k&=(1-\alpha_o)\widehat{\mathbf p}_{k-1}
 +\alpha_o\mathbf p_k,\\
\widehat{\dot{\mathbf q}}_k
 &=\frac{\widehat{\mathbf q}_k-\widehat{\mathbf q}_{k-1}}{T_c}.
\end{aligned}
\label{eq:policy_observer}
\end{equation}
Filtered states initialize at the first measurement, with zero initial
rate. A one-step joint jump exceeding $[20,10,20,30]$ degrees or a tip
jump exceeding 1,000 mm triggers an observer reset. A reset after
controller initialization is handled as a fault.

Define the pose encoding and policy-conditioning error as
\begin{equation}
\begin{aligned}
\mathbf s_k={}&[\sin(\kappa\widehat q_{1,k}),
\cos(\kappa\widehat q_{1,k}),
\widehat q_{2,k},\widehat q_{3,k},\widehat q_{4,k}]^{\mathsf T},\\
\boldsymbol\delta_k={}&\mathbf p^{\mathrm{pr}}_k-\widehat{\mathbf p}_k.
\end{aligned}
\end{equation}
The ordered observation is
\begin{equation}
\mathbf x_k=
\begin{bmatrix}
\mathbf s_k^{\mathsf T}&
\widehat{\dot{\mathbf q}}_k^{\mathsf T}&
\boldsymbol\delta_k^{\mathsf T}&
\|\boldsymbol\delta_k\|_2&h_k
\end{bmatrix}^{\mathsf T}
\in\mathbb R^{14},
\label{eq:policy_features}
\end{equation}
Here $h_k$ is a conditioning horizon in seconds, not a completion
deadline. The observation has no history stack or binary mask;
desired velocity enters Cartesian feedback separately.

Let $F_\Theta$ denote the fully connected
$14$--$256$--$256$--$128$--$4$ network with biases,
sigmoid linear unit (SiLU) hidden activations
$s(a)=a/(1+\exp(-a))$, and 103,044 trainable parameters.
It uses input normalization
$(\mathbf x-\boldsymbol\mu_x)\oslash\boldsymbol\sigma_x$
and a linear output followed by the fixed affine transformation
$(\mathbf A_y,\mathbf b_y)$. The anchored policy is
\begin{equation}
\begin{aligned}
\mathbf x_k^0&=
[\mathbf s_k^{\mathsf T},\mathbf0_4^{\mathsf T},
 \mathbf0_3^{\mathsf T},0,h_k]^{\mathsf T},\\
\mathbf u_{\mathrm{IL},k}
&=F_\Theta(\mathbf x_k)-F_\Theta(\mathbf x_k^0).
\end{aligned}
\label{eq:anchored_policy}
\end{equation}
Both evaluations share weights and normalization; the affine output
bias cancels. Outputs are swing, boom, arm, and bucket rates in
$\mathrm{deg\,s^{-1}}$. When
$\boldsymbol\delta_k=\mathbf0$ and
$\widehat{\dot{\mathbf q}}_k=\mathbf0$, nominal action is exactly
zero. Weights are trained offline and frozen online; command
limiting follows feedback and adaptation.

\subsection{Adaptive Cartesian Tracking Correction}
\label{subsec:robust_adaptive}

Feedback uses the current measured task error
$\mathbf e_k=\mathbf p_{d,k}-\mathbf p_k$, not the policy preview error.
Let $D_k=\|\mathbf e_k\|_2$, $\mathbf J_k=\mathbf J(\mathbf q_k)$,
and $\sigma_k=\sigma_{\min}(\mathbf J_k)$. Define
\begin{equation}
\begin{aligned}
\lambda_k&=1+8\max(0,1-\sigma_k/12)^2,\\
s_k&=\operatorname{clip}(\sigma_k/6,0.1,1),\\
\mathbf J_k^\#&=\mathbf J_k^{\mathsf T}
(\mathbf J_k\mathbf J_k^{\mathsf T}+\lambda_k^2\mathbf I_3)^{-1}.
\end{aligned}
\label{eq:damped_inverse}
\end{equation}
All numerical gains and thresholds below use the degree/millimetre
convention specified in Section~\ref{sec:problem}.

For compactness, define the causal filter
\begin{equation}
\mathcal L_\tau[\mathbf a]_k
=e^{-T_c/\tau}\mathcal L_\tau[\mathbf a]_{k-1}
 +(1-e^{-T_c/\tau})\mathbf a_k.
\end{equation}
All rate-filter and excitation-statistic states initialize at zero.
Both stored intervention flags initialize to false.
Measured Cartesian velocity is
$\overline{\mathbf v}_k=
\mathcal L_{0.25}[(\mathbf p_k-\mathbf p_{k-1})/T_c]$.
With integral state $\mathbf I_k$, the requested and limited Cartesian
feedback velocities are
\begin{equation}
\begin{aligned}
\mathbf w_k^*={}&k_{p,k}\mathbf e_k+\mathbf I_k
 +0.25(\mathbf v_{d,k}-\overline{\mathbf v}_k)+\rho_k\frac{\mathbf e_k}{\sqrt{D_k^2+25}},\\
\mathbf w_k={}&\frac{\mathbf w_k^*}
 {\max(1,\|\mathbf w_k^*\|_2/60)}, \hspace{0.1cm}
\mathbf u_{\mathrm{fb},k}={}s_k\mathbf J_k^\#\mathbf w_k.
\end{aligned}
\label{eq:cartesian_feedback}
\end{equation}

The gain/bias estimator compares measured motion with the nominal
position-filter prediction. Let $\tau_f=0.1$ s,
$a_f=\exp(-T_c/\tau_f)$, and let $\mathbf f_{1,k},\mathbf f_{2,k}$
be nominal filter states driven by the previously emitted demand:
\begin{equation}
\begin{aligned}
\mathbf f_{1,k}={}&\mathbf c_{k-1}
 +a_f(\mathbf f_{1,k-1}-\mathbf c_{k-1}),\\
\mathbf f_{2,k}={}&\mathbf c_{k-1}
 +a_f\big[(\mathbf f_{2,k-1}-\mathbf c_{k-1})\\
&\qquad+(T_c/\tau_f)(\mathbf f_{1,k-1}-\mathbf c_{k-1})\big].
\end{aligned}
\label{eq:nominal_filter_prediction}
\end{equation}
Both filter states initialize at $\mathbf c_0$.
The measured and predicted rate regressors are
\begin{equation}
\begin{aligned}
\mathbf z_k&=\mathcal L_{0.25}
 [(\mathbf q_k^{\mathrm{uw}}-\mathbf q_{k-1}^{\mathrm{uw}})/T_c],\\
\mathbf r_k&=\mathcal L_{0.25}
 [(\mathbf f_{2,k}-\mathbf f_{2,k-1})/T_c].
\end{aligned}
\end{equation}
With estimates carried from the previous sample, the residual is
\begin{equation}
\boldsymbol\varepsilon_k
=\mathbf z_k-(\widehat{\mathbf g}_{k-1}\odot\mathbf r_k
 +\widehat{\mathbf b}_{k-1}).
\label{eq:prediction_residual}
\end{equation}
These effective gain and bias estimates compensate deviations between
the measured joint rates and the nominal predicted response.

Excitation is assessed using
$m_{j,k}=\mathcal L_5[r_j]_k$,
$v_{j,k}=\mathcal L_5[r_j^2]_k$, and
$\nu_{j,k}=\max(0,v_{j,k}-m_{j,k}^2)$.
For an eligible joint, define
$\widetilde\varepsilon_{j,k}=
\operatorname{sgn}(\varepsilon_{j,k})
\max(|\varepsilon_{j,k}|-0.001,0)$ and update
\begin{equation}
\begin{aligned}
\widehat g_{j,k}
={}&\operatorname{clip}_{[0.35,1.8]}\Big\{
\widehat g_{j,k-1}+T_c\Big[
 \frac{0.8\widetilde\varepsilon_{j,k}r_{j,k}}
 {r_{j,k}^2+0.15^2}\\
&\hspace{31mm}-0.005(\widehat g_{j,k-1}-1)\Big]\Big\},\\
\widehat b_{j,k}
={}&\operatorname{clip}_{[-0.12,0.12]}\Big\{
\widehat b_{j,k-1}+T_c\Big[
 \frac{0.7\widetilde\varepsilon_{j,k}0.15^2}
 {r_{j,k}^2+0.15^2}\\
&\hspace{31mm}-0.005\widehat b_{j,k-1}\Big]\Big\}.
\end{aligned}
\label{eq:normalized_adaptation}
\end{equation}
The estimates initialize at $\widehat{\mathbf g}=\mathbf1$,
$\widehat{\mathbf b}=\mathbf0$. Updates require combined-control mode,
completion of the first 50 control samples, no intervention flag aligned
with the measured interval, $\nu_{j,k}>2.5\times10^{-5}$,
$|r_{j,k}|>0.005$, and $|\varepsilon_{j,k}|>0.001$.
Otherwise, the joint's estimates are held fixed, including their leakage
terms. The intervention flag is delayed to follow the emitted-demand
timing; with the register convention above it is the limiting flag from
two controller samples earlier.

When the previous sample had neither task-velocity nor governor limiting,
the scheduled gains are updated as
\begin{equation}
\begin{aligned}
k_{p,k}={}&\operatorname{clip}_{[0.2,0.6]}
\{k_{p,k-1}+T_c[k_{p,k}^{*}-k_{p,k-1}]\},\\
k_{p,k}^{*}={}&0.2+0.4D_k/(D_k+50),\\
\rho_k={}&\operatorname{clip}_{[0.5,10]}
\{\rho_{k-1}+0.7T_c[\rho_k^{*}-\rho_{k-1}]\},\\
\rho_k^{*}={}&\operatorname{clip}_{[0.5,10]}
 (0.5+\|\mathbf J_k\boldsymbol\varepsilon_k\|_2).
\end{aligned}
\end{equation}
Otherwise, they remain unchanged. Initial values are $k_{p,0}=0.2$
and $\rho_0=0.5$.

The combined and correction commands are
\begin{equation}
\begin{aligned}
\mathbf u_{\Sigma,k}
&=(\mathbf u_{\mathrm{IL},k}+\mathbf u_{\mathrm{fb},k}
 -\widehat{\mathbf b}_k)\oslash\widehat{\mathbf g}_k,\\
\mathbf u_{\mathrm{ad},k}
&=\mathbf u_{\Sigma,k}-\mathbf u_{\mathrm{IL},k}.
\end{aligned}
\label{eq:exact_command_composition}
\end{equation}
In IL-only mode, $\mathbf u_{\Sigma,k}=\mathbf u_{\mathrm{IL},k}$;
feedback, estimator, scheduled-gain, and integral updates are disabled.

After the governor determines $\mathbf u_k$, the combined mode uses
anti-windup integration. Let $\chi_k=1$ only when neither Cartesian
task limiting nor governor limiting occurs, and zero otherwise. Then
\begin{equation}
\begin{aligned}
\mathbf I_{k+1}=\Pi_{25}\Big\{\mathbf I_k+T_c\big[&
0.08\chi_k\mathbf e_k+(\mathbf w_k-\mathbf w_k^*)\\
&+\mathbf J_k\big(\widehat{\mathbf g}_k\odot
(\mathbf u_k-\mathbf u_{\Sigma,k})\big)\big]\Big\},
\end{aligned}
\label{eq:integral_antiwindup}
\end{equation}
$\Pi_{25}$ projects onto the Euclidean ball of radius 25 and
$\mathbf I_0=\mathbf0$.

\subsection{Shared Command Governor and Execution}
\label{subsec:governor}

Both modes share the same position-reference, speed, and acceleration
limits, in joint order swing, boom, arm, bucket:
\begin{equation}
\begin{aligned}
\mathbf q_{\min}&=[-\infty,-8,-172,-160]^{\mathsf T},\\
\mathbf q_{\max}&=[\infty,75,-22,60]^{\mathsf T},\\
\mathbf v_{\max}&=[0.6,0.4,0.6,0.8]^{\mathsf T},\\
\mathbf a_{\max}&=[0.6,0.5,0.8,1.0]^{\mathsf T}.
\end{aligned}
\end{equation}
Swing has no finite position bound in this model, but its rate and
acceleration are limited.

For speed $v\geq0$, define the discrete stopping distance
\begin{equation}
\mathcal D_j(v)=T_c\sum_{n=0}^{\infty}
\max(v-na_{\max,j}T_c,0).
\end{equation}
Only finitely many summands are nonzero. The admissible stopping speed is
\begin{equation}
\mathcal S_j(d)=\max\{v\in[0,v_{\max,j}]:
\mathcal D_j(v)\leq\max(d,0)\},
\end{equation}
with $\mathcal S_j(\infty)=v_{\max,j}$.
The governor uses the position-demand register, rather than measured
plant position, to form
\begin{equation}
\begin{aligned}
\ell_{j,k}=\max\{&-\mathcal S_j(c_{j,k}-q_{\min,j}),
u_{j,k-1}-a_{\max,j}T_c\},\\
b_{j,k}^{+}=\min\{&\mathcal S_j(q_{\max,j}-c_{j,k}),
u_{j,k-1}+a_{\max,j}T_c\}.
\end{aligned}
\end{equation}
Let $\zeta_k=1$ during fixed-goal regulation when the filtered
goal distance is at most $25\,\mathrm{mm}$, and $\zeta_k=0$
otherwise. For timed tracking, $\zeta_k=0$.
For IL-only and IL-ACT, the governor input is
\begin{equation}
\mathbf u_{\mathrm{req},k}
=(1-\zeta_k)\mathbf u_{\Sigma,k}.
\label{eq:goal_stopping_request}
\end{equation}
Thus, stopping requests zero rate upstream of the acceleration
and stopping-distance constraints.
A governor sample is feasible when $\mathbf c_k$ is finite, lies within
the declared position envelope, and $\ell_{j,k}\leq b_{j,k}^{+}$ for
every joint. On a feasible sample,
\begin{equation}
u_{j,k}=\operatorname{clip}
(u_{\mathrm{req},j,k},\ell_{j,k},b_{j,k}^{+}),
\label{eq:simultaneous_supervisor}
\end{equation}
followed by the reference update in
\eqref{eq:reference_integrator}. Singularity scaling is applied to the
Cartesian feedback in \eqref{eq:cartesian_feedback}, not to the full
nominal-plus-correction command in this governor.
Invalid or nonfinite inputs, an invalid observer, inconsistent startup,
or a subsequent observer reset trigger a latched fault. A fault requests
zero rate before the governor, so the normal response is limited
deceleration rather than an instantaneous stop. If the admissible
interval itself is infeasible, the fallback is
\begin{equation}
u_{j,k}=\operatorname{sgn}(u_{j,k-1})
\max(|u_{j,k-1}|-a_{\max,j}T_c,0),
\end{equation}
with an infeasibility fault. The fallback reduces the previous
joint-rate command toward zero using the deceleration.
Reference admissibility is established under the feasibility
conditions stated below.
The initialization sample holds the initial demand and emits zero rate.

Each valid sample updates observations and prediction statistics
before eligible estimation, gain scheduling, and command composition.
Stopping and the governor precede anti-windup integration.
The controller emits $\mathbf c_k$ and stores $\mathbf c_{k+1}$
with interval-aligned histories. IL-only omits feedback and adaptation.

\subsection{Controller Boundedness and Reference Admissibility}
\label{subsec:controller_properties}

Projection and scheduling ensure
$0.35\leq\widehat g_{j,k}\leq1.8$,
$|\widehat b_{j,k}|\leq0.12$,
$0.2\leq k_{p,k}\leq0.6$,
$0.5\leq\rho_k\leq10$, and $\|\mathbf I_k\|_2\leq25$.
Furthermore, $\lambda_k\geq1$ and
$\|\mathbf J_k^\#\|_2\leq(2\lambda_k)^{-1}$, while
$\|\mathbf w_k\|_2\leq60$. Therefore the feedback command is bounded.
If the learned nominal output is bounded on the operating domain, the
combined command is also bounded because the estimated gains remain
strictly positive.

At each feasible governor sample,
\begin{equation}
\begin{aligned}
|u_{j,k}|&\leq v_{\max,j}, \hspace{0.1cm} |u_{j,k}-u_{j,k-1}|\leq a_{\max,j}T_c,\\
q_{\min,j}&\leq c_{j,k}+T_cu_{j,k}\leq q_{\max,j}.
\end{aligned}
\end{equation}
The position inequality follows because the discrete stopping distance
contains the next displacement $T_c|u_{j,k}|$.
Consequently, an initially admissible position-demand register remains
admissible while the governor remains feasible.

The analysis establishes bounded adaptive states and Cartesian feedback commands and admissible
generated references under the stated feasibility conditions.
Closed-loop task performance is evaluated in
Section~\ref{sec:evaluation}.

\section{Simscape Evaluation}
\label{sec:evaluation}

\subsection{Original Training and Analytical Rollouts}
The original single-seed policy passes all 32 validation cases
at each of four stages; mean terminal error falls from $21.431$
to $20.367$ mm. It passes $63/64$ separate analytical tests,
with all-case mean error $26.519$ mm, including the $397.481$ mm
timeout. These rollouts use the nominal response and governor.
Success requires $25$ mm position and
$0.05\,\mathrm{deg\,s^{-1}}$ speed tolerances, checked every
$0.01$ s, with $1$ s qualification, $3$ s hold, and $600$ s
timeout. Training curves and casewise results are in Sec. S2;
the additional three-seed study follows in
Section~\ref{subsec:expanded_tracking}.

\subsection{Original Goal and Spiral Benchmarks}
The control comparison uses four-joint Cartesian goal regulation and
timed trajectory tracking. The IL-only mode and the combined mode
are defined in Section~\ref{sec:control}. For sequential goals, the
current goal is held until the qualification and hold conditions are
satisfied, with a $600$ s timeout. Policy, observer, reference-register,
and plant states continue across goal changes. For timed tracking,
current reference position and velocity advance every $0.1$ s;
intermediate target waiting and error-dependent time dilation are absent.

The policy-conditioning horizon is $h_k=2\,\mathrm{s}$ in the
original goal-regulation and timed-tracking benchmarks. For goal regulation,
$\mathbf p^{\mathrm{pr}}_k=\mathbf p_{d,k}$ is the current goal.
For timed tracking,
\begin{equation}
\mathbf p^{\mathrm{pr}}_k=\mathbf p_d(t_k+2\,\mathrm{s}),\qquad
\mathbf v_{d,k}=\dot{\mathbf p}_d(t_k),
\label{eq:experimental_preview}
\end{equation}
with Cartesian feedback using $\mathbf p_d(t_k)$. Preview evaluation
includes transitions between the approach, spiral, and terminal hold.
Define $H(\xi)=35\xi^4-84\xi^5+70\xi^6-20\xi^7$ for
$0\leq\xi\leq1$. For $0\leq t<600$ s, the reference is
$\mathbf p_d(t)=\mathbf f(\mathbf q_0+
H(t/600)(\mathbf q_s-\mathbf q_0))$, where $\mathbf f$ is the
forward kinematics in \eqref{eq:cartesian_task},
$\mathbf q_0=[0,30,-100,-20]^{\mathsf T}$ degrees, and
$\mathbf q_s$ is the elbow-down inverse-kinematic solution for
$[9750,0,4800]^{\mathsf T}$ mm with zero swing and cumulative
bucket pitch $q_{2,s}+q_{3,s}+q_{4,s}=16.5^\circ$.
Let $T_{\mathrm{sp}}$ denote the baseline spiral tracking duration.
For $600\,\mathrm{s}\leq t<600\,\mathrm{s}+T_{\mathrm{sp}}$, set
$s=H((t-600\,\mathrm{s})/T_{\mathrm{sp}})$ and
$R(s)=3750/(1+2s)$. The two-revolution spiral is
\begin{equation}
\mathbf p_d(t)=
\begin{bmatrix}
6000-1400s^2+R(s)\cos(4\pi s)\\
R(s)\sin(4\pi s)\\
4800-6500[1-(1-s)^4]
\end{bmatrix}\,\mathrm{mm}.
\label{eq:experimental_spiral}
\end{equation}
For $t\geq600\,\mathrm{s}+T_{\mathrm{sp}}$,
$\mathbf p_d(t)=[5850,0,-1700]^{\mathsf T}$ mm
and $\mathbf v_d(t)=\mathbf0$. Desired velocity is obtained
analytically from the position reference. The run ends after
a $30$ s terminal hold following completion of the spiral.

\begin{figure}[tb]
\hspace*{-0.0cm} 
\centering
\scalebox{0.85}{\includegraphics[trim={0cm 0.0cm 0.0cm 0cm},clip,width=\columnwidth]{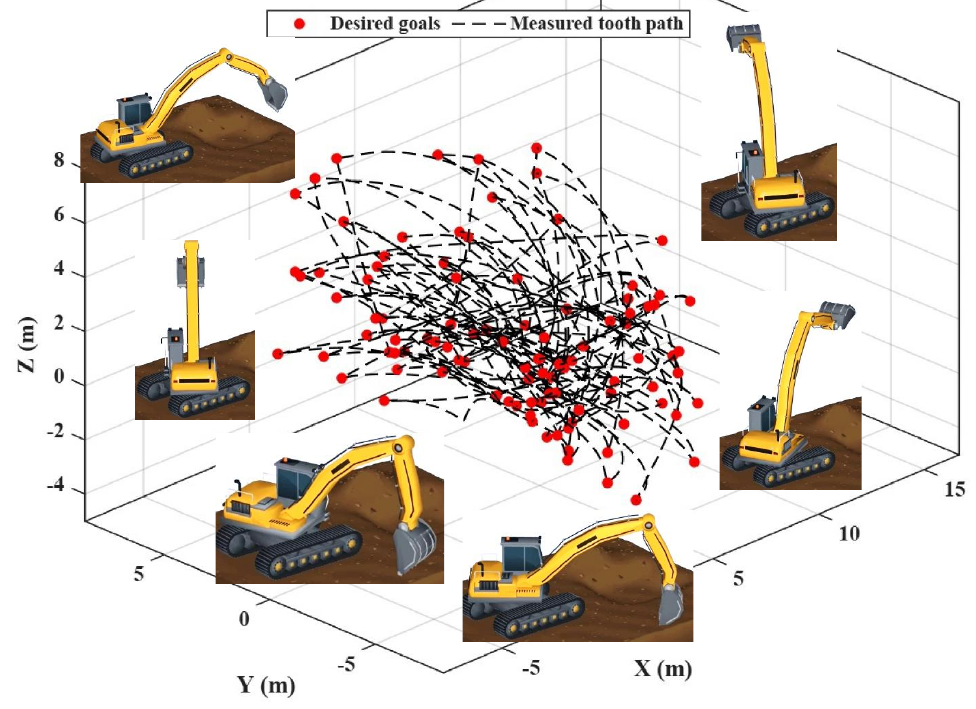}}
\caption{IL-ACT regulation to 100 Cartesian goals with the disturbance.}
\label{goal_reach}
\end{figure}

\begin{figure}[tb]
\hspace*{-0.0cm} 
\centering
\scalebox{0.85}{\includegraphics[trim={0cm 0.0cm 0.0cm 0cm},clip,width=\columnwidth]{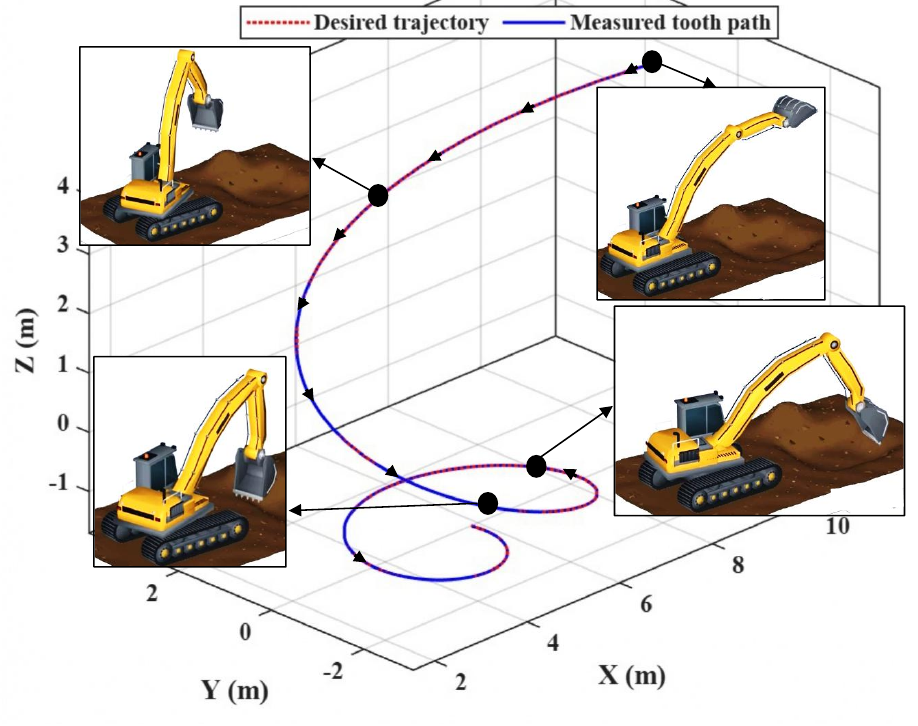}}
\caption{IL-ACT spiral tracking with the hydraulic disturbance.}
\label{3dsipral}
\end{figure}

The added response qualitatively represents selected hydraulic
effects discussed in~\cite{RudermanHydraulic2017,sadsadshahna2025robust}.
It updates at $0.01$ s and includes valve/velocity lag, dead
zone, directional gains, friction/release hysteresis, and supply/load
modulation before two nominal $0.1$ s position-input filters.
Disturbed runs share the response law and absolute-time schedule;
realized load and exposure depend on configuration and completion
time. Equations, initial states, and parameters appear in Sec. S3.
In these original benchmarks, IL-only and IL-ACT use identical
learned policy parameters and joint-command bounds. Goal-regulation runs share the ordered set
of 100 goals, with performance-dependent transition times; spiral
runs share the timed reference above. All eight original IL-only and IL-ACT runs recorded zero controller-fault
samples and passed the recorded joint-limit checks. Their four
goal-regulation runs completed without timeouts, and their four
spiral runs remained within $25$ mm over the evaluated spiral interval. Gain/bias updates occurred only
in the disturbed IL-ACT runs, with $11192$ joint-update events
for goal regulation and $10014$ for spiral tracking.
Figures~\ref{goal_reach} and~\ref{3dsipral} illustrate IL-ACT
bucket-tip motion during goal regulation and spiral tracking,
respectively. Commanded and measured joint traces are provided in Sec. S5.
All controllers share the stopping, qualification, hold, and governor
rules in Section~\ref{subsec:governor}. Joint command-tracking RMSE
uses $q_{j,k}-c_{j,k}$ on the complete $0.1$ s grid, with swing
differences wrapped to $[-180,180)$ degrees. The largest joint
RMSE is the maximum across joints.
Jointwise corrective-rate and combined stopping/governor-modification
statistics, including their definitions and evaluation intervals,
are reported in Sec. S5.

For the spiral results in
Table~\ref{tab:excavator_control_comparison_spiral}, Cartesian tracking
error is the Euclidean distance between the measured and desired
bucket-tip positions. Its root-mean-square (RMS) value and mean are computed
using trapezoidal time integration, including both
endpoints. The 95th percentile uses linear interpolation of the sorted
error samples. Full-run and terminal statistics are reported separately.

\begin{table}[!tbp]
\centering
\caption{Goal regulation over 100 attempts.}
\label{tab:excavator_control_comparison}

\footnotesize
\setlength{\tabcolsep}{3pt}
\renewcommand{\arraystretch}{1.10}

\begin{tabularx}{\linewidth}{
    @{}>{\raggedright\arraybackslash}Xrrr@{}
}
\toprule
\multicolumn{4}{c}{\textbf{No added disturbance}} \\
\midrule
Metric & PID & IL-only & IL-ACT \\
\midrule
Independently verified goals
& 95/100 & 100/100 & 100/100 \\
Total recorded duration (s)
& 16000 & 14370.9 & 13682.2 \\
Median goal-segment duration (s)
& 145 & 130.20 & 126.45 \\
Maximum goal-segment duration (s)
& 600 & 293.5 & 287.5 \\
Mean terminal Cartesian error (mm)
& 22 & 20.095 & 4.446 \\
Maximum terminal Cartesian error (mm)
& 24 & 21.238 & 11.015 \\
Maximum error during accepted holds (mm)
& 24.5 & 21.238 & 11.531 \\
Largest joint command-tracking RMSE (deg)
& 0.15 & 0.132828 & 0.139972 \\
Gain/bias joint-update events
& 0 & 0 & 0 \\
\midrule
\multicolumn{4}{c}{\textbf{Response disturbance}} \\
\midrule
Metric & PID & IL-only & IL-ACT \\
\midrule
Independently verified goals
& 86/100 & 100/100 & 100/100 \\
Total recorded duration (s)
& 19000 & 14506.8 & 13698.0 \\
Median goal-segment duration (s)
& 175 & 132.60 & 126.35 \\
Maximum goal-segment duration (s)
& 600 & 296.1 & 290.9 \\
Mean terminal Cartesian error (mm)
& 23 & 1.125 & 4.690 \\
Maximum terminal Cartesian error (mm)
& 24.8 & 2.596 & 11.118 \\
Maximum error during accepted holds (mm)
& 24.9 & 11.860 & 11.202 \\
Largest joint command-tracking RMSE (deg)
& 0.23 & 0.184747 & 0.195247 \\
Gain/bias joint-update events
& 0 & 0 & 11192 \\
\bottomrule
\end{tabularx}
\end{table}


The PID baseline operates in Cartesian velocity space, with
desired-velocity feedforward, proportional and integral position-error
feedback, and filtered velocity-error feedback. Its request is limited
to $60\,\mathrm{mm\,s^{-1}}$ and mapped through the damped inverse
and singularity scaling in \eqref{eq:damped_inverse}. It uses
back-calculation anti-windup and the same command governor, emitted-demand
timing, and goal-acceptance rules as the IL-based controllers.
Nine diagonal gains are tuned using JAYA~\cite{shahna2024robustDSADAS}
on separate cases excluding the evaluation goals, then frozen for all
nominal and disturbed evaluations. Three seeded runs use 3060 objective
evaluations, with equal weighting of nominal/disturbed goal and tracking
groups. The full PID law, gains, tuning cases, objective, and failure
handling are given in Sec. S4.


\begin{table}[!tbp]
\centering
\caption{PID, IL-only, and IL-ACT spiral tracking.}
\label{tab:excavator_control_comparison_spiral}

\footnotesize
\setlength{\tabcolsep}{3pt}
\renewcommand{\arraystretch}{1.10}

\begin{tabularx}{\linewidth}{
    @{}>{\raggedright\arraybackslash}Xrrr@{}
}
\toprule
\multicolumn{4}{c}{\textbf{No added disturbance}} \\
\midrule
Metric & PID & IL-only & IL-ACT \\
\midrule
Cartesian tracking RMSE (mm)
& 5.72946 & 2.49800 & 0.04459 \\
Mean Cartesian tracking error (mm)
& 3.51830 & 1.73826 & 0.02830 \\
95th-percentile tracking error (mm)
& 10.56082 & 4.85366 & 0.09465 \\
Maximum tracking error (mm)
& 18.94713 & 8.16079 & 0.18485 \\
Maximum error over full run (mm)
& 45.31658 & 26.74581 & 0.81243 \\
Final error after terminal hold (mm)
& 0.001
& $5.069\!\times\!10^{-5}$
& $9.218\!\times\!10^{-6}$ \\
Gain/bias joint-update events
& 0 & 0 & 0 \\
\midrule
\multicolumn{4}{c}{\textbf{Response disturbance}} \\
\midrule
Metric & PID & IL-only & IL-ACT \\
\midrule
Cartesian tracking RMSE (mm)
& 12.48371 & 2.49815 & 0.05864 \\
Mean Cartesian tracking error (mm)
& 8.92604 & 1.74221 & 0.04058 \\
95th-percentile tracking error (mm)
& 25.15782 & 4.85496 & 0.13607 \\
Maximum tracking error (mm)
& 45.64039 & 8.19303 & 0.40440 \\
Maximum error over full run (mm)
& 90.81526 & 26.72286 & 0.88102 \\
Final error at $t=7630$ s (mm)
& 0.23029 & 0.00608 & 0.06451 \\
Gain/bias joint-update events
& 0 & 0 & 10014 \\
\bottomrule
\end{tabularx}
\end{table}

Teacher+ACT replaces only the learned nominal command with the
teacher (Sec. S2), retaining observations, ACT, and acceptance rules.
Both complete all 100 goals. Under nominal and disturbed response,
IL-ACT reduces duration by $7.22\%$ and $12.97\%$, and mean
terminal error by approximately $16.16\%$ and $28.39\%$,
respectively. Averaging the original spiral per-condition metrics
gives $27.88\%$ lower RMSE but $33.18\%$ higher maximum error.
Enabling estimation lowers disturbed-goal mean terminal error
by $12.68\%$ (Sec. S5). Full original comparisons are in Sec. S5.

\subsection{Expanded Tracking and Initialization Study}
\label{subsec:expanded_tracking}
\begin{figure}[t]
\centering
\begin{minipage}[b]{0.49\columnwidth}
\centering
\includegraphics[width=\linewidth]{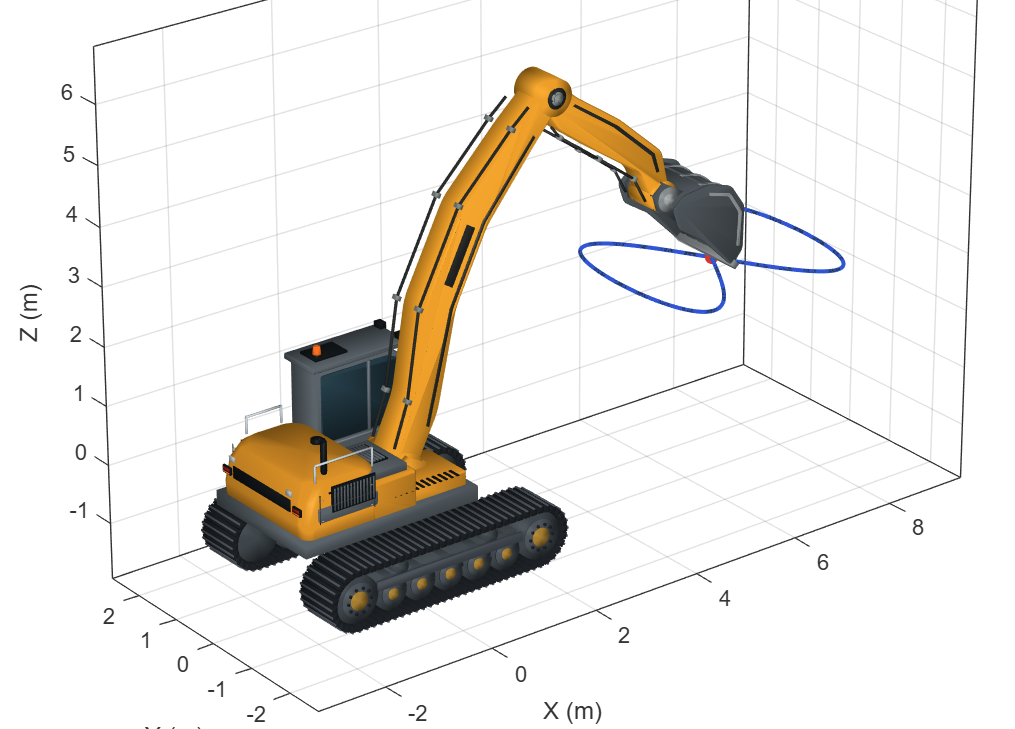}
\par\smallskip
{\footnotesize (a) Figure-eight}
\end{minipage}\hfill
\begin{minipage}[b]{0.49\columnwidth}
\centering
\includegraphics[width=\linewidth]{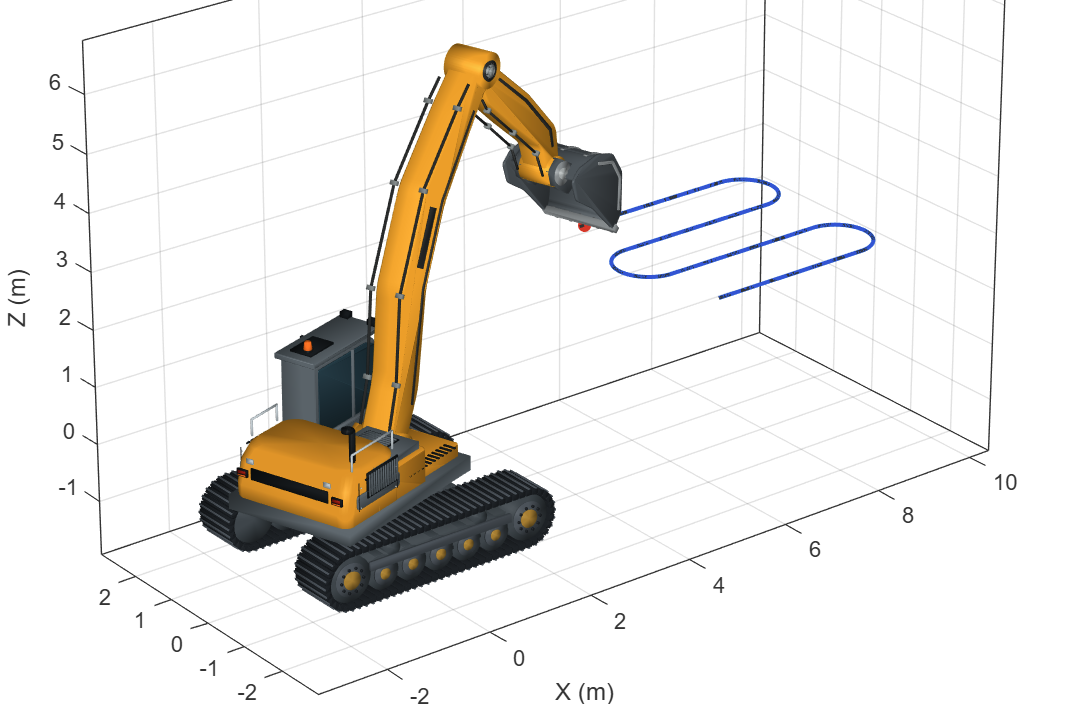}
\par\smallskip
{\footnotesize (b) Rounded raster}
\end{minipage}
\caption{Additional evaluation-path visualizations.}
\label{fig:additional_paths}
\end{figure}

For each seed $s\in\{11,22,33\}$, telemetry- and
random-initialized policies receive $64{,}920$ AdamW updates
with replacement-sampled batches of $1{,}024$ examples.
Each pair shares initial and pooled aggregation data, minibatch
indices, normalization, and optimizer settings. This equal
refinement budget excludes telemetry pretraining; all
final-stage checkpoints are evaluated.
The figure-eight and rounded-raster paths
(Fig.~\ref{fig:additional_paths}) are excluded from training,
tuning, and checkpoint selection. Runs use a $600$ s approach
and $30$ s terminal observation. Each path is evaluated at
temporal rate factors $\gamma\in\{1,2\}$, corresponding to the
$1\times$ and $2\times$ conditions, with tracking duration
$T_{\mathrm{sp}}/\gamma$ and the progress polynomial $H$
defined above. The $1\times$ condition uses the baseline
spiral tracking duration; the $2\times$ condition traverses
the same geometric path in half that duration.
The noisy and additional-load spiral conditions use $\gamma=1$.
Controller/reference clocks advance every $0.1$ s independently
of error. Nominal mappings use a shared $2$ s preview; ACT uses
current position/velocity references. Controller settings in
Section~\ref{sec:control} are shared, except for the intended
frozen/updating estimator ablation.
Ten conditions comprise both new paths at both speeds under
nominal/hydraulic response, plus noisy and additional-load
spirals. Sensing adds independent zero-mean Gaussian noise to
all four joint angles ($\sigma_q=0.02236^\circ$, $10$ Hz); Cartesian measurements
use FK of the same corrupted angles. All comparators share one
realization (seed \texttt{2026091201}) with nominal actuator
response. The load condition scales both directional hydraulic
gains by $\sqrt{0.5}$, retaining all other response parameters.
Per-run RMSE is
$\sqrt{N^{-1}\sum_k\|\mathbf p_{\mathrm{true}}(t_k)
-\mathbf p_{\mathrm{ref}}(t_k)\|_2^2}$, using uncorrupted simulated
position and equally weighted $0.1$ s samples. Tracking windows
are $[600\,\mathrm{s},600\,\mathrm{s}+T_{\mathrm{sp}}/\gamma]$,
with $N=1+T_{\mathrm{sp}}/(\gamma T_c)$ samples and both
endpoints included. Here $\gamma=1$ for the baseline and
spiral conditions, and $\gamma=2$ for the doubled temporal-rate
conditions.
Table~\ref{tab:expanded_tracking} averages the three policy-seed
RMSEs. Additional protocol details are in Sec. S6.

\begin{table}[!tbp]
\centering
\caption{Cartesian tracking RMSE (mm). Learned entries
average seeds $11$, $22$, and $33$.
N/H denote nominal/hydraulic-response conditions.}
\label{tab:expanded_tracking}
\footnotesize
\setlength{\tabcolsep}{3pt}
\begin{tabularx}{\columnwidth}{@{}Xrrr@{}}
\toprule
Path, speed, condition & Teacher & Telemetry & Random\\
\midrule
Figure-eight, $1\times$, N & 0.04233 & 0.01414 & 0.01286\\
Figure-eight, $1\times$, H & 0.04897 & 0.03121 & 0.02973\\
Figure-eight, $2\times$, N & 0.16931 & 0.05663 & 0.05828\\
Figure-eight, $2\times$, H & 0.17196 & 0.06428 & 0.06553\\
Rounded raster, $1\times$, N & 0.07278 & 0.03194 & 0.03498\\
Rounded raster, $1\times$, H & 0.07851 & 0.04217 & 0.04362\\
Rounded raster, $2\times$, N & 0.28851 & 0.13423 & 0.14785\\
Rounded raster, $2\times$, H & 0.28847 & 0.13581 & 0.14878\\
Spiral, sensor noise & 3.95121 & 2.85776 & 3.81691\\
Spiral, additional load & 0.07758 & 0.05506 & 0.05720\\
\bottomrule
\end{tabularx}
\end{table}

Telemetry-initialized IL-ACT lowers RMSE versus Teacher+ACT in
all 24 figure-eight and rounded-raster seed comparisons and all
three additional-load spiral comparisons. The nine non-noise condition means improve by approximately
$29$--$67\%$. Under the shared sensor-noise realization,
telemetry-initialized IL-ACT achieves $27.67\%$ lower mean
RMSE than Teacher+ACT.
Telemetry initialization improves $16/30$ comparisons against
random weights under shared telemetry-derived normalization,
pooled data, and matched refinement, showing no consistent benefit.
With fixed policies, estimation reduces mean noise RMSE by
$22.44\%$ and mean per-run tracking maximum by $17.5\%$
(Table~\ref{tab:noise_estimator}), improving both for every seed
under one shared noise realization.

\begin{table}[!tbp]
\centering
\caption{Sensor-noise estimator ablation; errors in mm.}
\label{tab:noise_estimator}
\footnotesize
\setlength{\tabcolsep}{4pt}
\begin{tabularx}{\columnwidth}{@{}Xrr@{}}
\toprule
Metric (telemetry-initialized
IL-ACT) & Frozen & Updating\\
\midrule
Mean of per-run RMSEs & 3.68452 & 2.85776\\
Mean of per-run tracking maxima & 12.086 & 9.974\\
\bottomrule
\end{tabularx}
\end{table}

This includes estimator-dependent gain scheduling; results
outside noise are mixed in Sec. S7. All $88$ runs completed with no recorded controller faults
or operating-limit violations; all matching checks passed.
The largest tracking-window error was $12.515$ mm
($25$ mm criterion).


\section{Discussion and Conclusions}
IL-ACT improves goal-regulation duration and spiral RMSE over
IL-only, and goal duration and terminal accuracy over Teacher+ACT.
Lower RMSE versus Teacher+ACT extends to figure-eight and
rounded-raster paths at both speeds and response settings,
and additional-load spiral tracking. Estimator benefits depend
on conditions; pretrained-weight effects remain mixed.
Tradeoffs include higher joint command-tracking RMSE and
disturbed-goal terminal error than IL-only. Analysis establishes
bounded adaptive states and Cartesian feedback, with reference
admissibility conditional on governor feasibility.
These findings concern the evaluated simulation conditions.

\bibliographystyle{ieeetr}
\bibliography{lcsys}

\end{document}